\documentclass[11pt]{article}

\usepackage[utf8]{inputenc}       
\usepackage[T1]{fontenc}          
\usepackage{graphicx}             
\usepackage{amsmath, amssymb}     
\usepackage{booktabs}             
\usepackage{multirow}             
\usepackage{array}                
\usepackage{url}                  
\usepackage[small]{caption}       
\usepackage{threeparttable}       
\usepackage{geometry}             
\usepackage{setspace}             
\usepackage{algpseudocode}        
\usepackage{algorithm}            
\usepackage[numbers]{natbib}      
\usepackage{hyperref}             

\hypersetup{
    colorlinks=true,
    linkcolor=blue,
    citecolor=blue,
    urlcolor=blue,
    pdftitle={Your Paper Title},
    pdfauthor={Your Name},
    pdfsubject={Your Subject},
    pdfkeywords={Keyword1, Keyword2, Keyword3}
}

\date{}

\title{Detection and Analysis of Offensive Online Content in Hausa Language} 

\author{
Fatima Muhammad Adam$^1$\and
Abubakar Yakubu Zandam$^2$\and
Isa Inuwa-Dutse$^3$ \\ 
$^1$Federal University Dutse, Nigeria\\
$^2$ Federal University of Technology Babura, Nigeria\\
$^3$University of Huddersfield, UK\\
\{fateemamohdadam2, ayzandam95\}@gmail.com,i.inuwa-dutse@hud.ac.uk}

\begin{document}
\maketitle

\begin{abstract}
Hausa, a major Chadic language spoken by over 100 million people mostly in West Africa is considered a low-resource language from a computational linguistic perspective. This classification indicates a scarcity of linguistic resources and tools necessary for handling various natural language processing (NLP) tasks, including the detection of offensive content. To address this gap, we conducted two set of studies (1) a user study ($n=101$) to explore cyberbullying in Hausa and (2) an empirical study that led to the creation of the first dataset of offensive terms in the Hausa language. We developed detection systems trained on this dataset and compared their performance against relevant multilingual models, including Google Translate. Our detection system successfully identified over 70\% of offensive, whereas baseline models frequently mistranslated such terms. We attribute this discrepancy to the nuanced nature of the Hausa language and the reliance of baseline models on direct or literal translation due to limited data to build purposive detection systems. These findings highlight the importance of incorporating cultural context and linguistic nuances when developing NLP models for low-resource languages such as Hausa. A post hoc analysis further revealed that offensive language is particularly prevalent in discussions related to religion and politics. To foster a safer online environment, we recommend involving diverse stakeholders with expertise in local contexts and demographics. Their insights will be crucial in developing more accurate detection systems and targeted moderation strategies that align with cultural sensitivities. 
\vspace{2mm}

\textbf{Trigger Warning: Readers may find some of the terms in this study distressing or disturbing; all examples are for illustration only.} 
\end{abstract}


\section{Introduction}
\label{sec:introduction} 
The modern world is highly interconnected, with millions of people and devices linked through the Internet. This interconnectivity has facilitated research and communication, driving transformative changes across various domains ~\cite{kennedy2006beyond,warf2018sage}. One significant outcome of this connectivity is the rise of online social networks, which enable user interactions and generate vast datasets for research purposes \cite{miller2010data,kumar2014twitter,haffner2018spatial}. Social media platforms such as Facebook\footnote{\url{https://facebook.com}} and Twitter (now X\footnote{\url{https://twitter.com}}) allow users to stay connected, express opinions, organize civil movements, and more \cite{kumar2014twitter,rane2012social,olteanu2015characterizing,freelon2016beyond,puspitasari2017arab,nemes2021information,dambo2022office,bello2023endsars}. These platforms have a profound impact on politics, governance, the economy, business, and other societal issues \cite{dollarhide2021social}.

However, online participation often involves the use of offensive, racist, and sexist language, contributing to the growing problem of cyberbullying \cite{caselli2020feel,caselli2020hatebert}. Cyberbullying has severe social and psychological consequences, including an increased risk of suicide, particularly among young people \cite{bullystat,magee2013teens}. In response to this issue, various strategies have been proposed to detect, mitigate, and combat cyberbullying \cite{zhong2016content,altay2018detection,zois2018optimal,rafiq2018scalable,yao2019cyberbullying,amin2020various,plaza2021multi}. However, these detection methods and resources are primarily designed for high-resource languages, rendering them ineffective for low-resource languages such as Hausa.

This study aims to support efforts to combat the proliferation of offensive content in low-resource languages, with a specific focus on Hausa. To achieve this, we outline the following objectives: 

    \begin{itemize}
        \item[-] Conduct a comprehensive review of existing research on offensive content detection in low-resource languages, with an emphasis on Hausa.
        \item[-] Engage native speakers through a user study to develop a robust dataset of offensive content in the Hausa language.
        \item[-] Develop an effective detection system tailored to identifying offensive content in Hausa.
        \item[-] Provide actionable recommendations and strategies to mitigate the spread of offensive content in Hausa.
    \end{itemize}
        
Beyond these core objectives, our study will also (1) investigate the prevalence of offensive language in Hausa and differentiate between banter and more harmful offensive terms (2) examine the interplay between idiomatic expressions with subtle abusive undertones, and (3) identify the topical issues most likely to attract offensive content. We believe this multi-faceted approach will provide a comprehensive understanding of offensive content in Hausa, leading to more effective detection and intervention strategies. 
As one of the first studies on the detection of offensive online content in the Hausa language, our contributions include:
    \begin{itemize}
        \item[-] First annotated dataset on offensive content in Hausa: We introduce the first collection of annotated datasets on offensive content in Hausa (HOC), enriching resources for future research and the development of online safety tools.
        \item[-] Hausa offensive content detection model: We develop a robust detection model specifically designed to identify offensive language in Hausa online discourse.
        \item[-] Insights into Hausa cyberbullying: Through user studies, we provide insights into the prevalence and nature of cyberbullying in Hausa, along with mitigation strategies to foster more civil online interactions.
    \end{itemize}

The remainder of this paper is structured as follows. Section~\ref{sec:related-work} reviews relevant literature and background studies. Section~\ref{sec:methodology} describes our approach, including data collection methods. Section~\ref{sec:experiment} presents the implementation details, while Section~\ref{sec:results-discussion} discusses the results and key findings. Finally, Section~\ref{sec:conclusion} concludes the study and outlines future research directions.


\section{Background and Related Work}
\label{sec:related-work}
In this section, we review relevant studies on hate speech, abusive content, and downstream tasks in the Hausa language.

\subsection{Hate Speech Detection}
According to the General Policy Recommendation of the European Commission against Racism and Intolerance (ECRI), hate speech encompasses any form of advocacy, promotion, or incitement, in any form, of disparagement, abhorrence, criticism, harassment, negative labelling, stigmatization, threats, insults, or other harmful expressions directed at individuals or groups \cite{ECRI2023}. Offensive language is closely related to various linguistic and societal issues, including abusive and violent tone, cyberbullying, racism, extremism, radicalization, toxicity, profanity, flaming, discrimination, and hateful speech \cite{caselli2020feel,founta2018large}. With the massive increase in social media interactions, offensive content and other forms of cyberbullying have also risen. In May 2016, the EU Commission reached an agreement with major technology companies\footnote{Facebook, Microsoft, Twitter, and YouTube} on a Code of Conduct to counter illegal hate speech online \cite{eucomm2016}. These measures have proven most effective in high-resource languages such as English and French. However, detecting cyberbullying-related content in low-resource languages like Hausa remains challenging due to the lack of annotated datasets and relevant linguistic resources. This study aims to address these challenges. 

Several studies based on tweets from X have been employed to detect hateful and abusive content across various social issues \cite{sanguinetti2018italian,basile2019semeval,mathew2021hatexplain}. For instance, tweets related to hate speech against immigrants \cite{sanguinetti2018italian} and misogynistic content in Spanish and English \cite{basile2019semeval} have been utilized to develop multilingual detection systems. Additionally, memes have been incorporated into multimodal classification systems to enhance hate speech detection by leveraging contextual understanding \cite{kiela2020hateful}. A benchmark dataset, HateXplain, consisting of hate, offensive, and neutral data points for detecting hate speech, has been introduced by \cite{mathew2021hatexplain}. 
Inherent racial biases in abusive language detection systems remain a significant concern, as discrepancies in accent and writing style can lead to discriminatory outcomes. Studies have highlighted that such biases disproportionately affect African-American users, undermining detection efforts intended to protect communities that are frequently targeted \cite{davidson2019racial,davidson2019racial}.

\subsection{Offensive Content Detection}
The Oxford English Dictionary defines offensive content as \textit{highly unpleasant and insulting; characterized by persistent violence and cruelty} \cite{dictionary_key}. As noted earlier, cyberbullying manifests in various forms, necessitating diverse detection strategies \cite{zhong2016content,al2016cybercrime,altay2018detection,van2018automatic,zois2018optimal,rafiq2018scalable,yao2019cyberbullying}. 
For instance, \cite{zhong2016content} focused on identifying forms of cyberbullying associated with image data. Another study applied a model that categorizes posts written by bullies, victims, and bystanders to detect online bullying \cite{van2018automatic}. Other notable detection strategies emphasize two main aspects: scalability and timeliness of detection systems. Timeliness is a crucial factor in the detection pipeline, ensuring efficient support for cyberbullying victims. A multi-stage cyberbullying detection system incorporating a scheduling mechanism has been proposed to enhance detection efficiency \cite{rafiq2018scalable}. Likewise, a sequential hypothesis testing approach has been suggested as a solution to scalability and timeliness challenges \cite{zois2018optimal}. \cite{yao2019cyberbullying} explored ways to reduce false positives while improving scalability and timeliness, using data from Instagram. 
Given that offensive content detection and sentiment analysis are closely related \cite{schmidt2017survey,oriola2020evaluating}, sentiment analysis techniques have been employed to identify offensive and inappropriate online content \cite{plaza2021multi}. Notably, the studies mentioned above primarily focus on cyberbullying detection in high-resource languages. However, there is growing interest in addressing offensive content and hate speech in low-resource languages such as Tamil \cite{rajalakshmi2023hottest,balakrishnan2023tamil}, Pashto \cite{khan2023offensive}, Urdu \cite{saeed2023detection}, and Persian \cite{kebriaei2023persian}. Efforts have also been made to enhance resources for tackling offensive and hateful content detection in these languages \cite{kovacs2022leveraging,sinyangwe2023detecting,miao2023detecting,markov2023holistic}.

\subsection{Downstream Tasks in the Hausa Language}
Datasets from online social media platforms have been leveraged for various computational linguistic tasks, including sentiment analysis \cite{zishumba2019sentiment,sanisentiment}, topic modelling \cite{kusumawardani2017topic,antypas2022twitter}, sexist language detection \cite{aliyu2023hausanlp}, hate speech recognition \cite{machuve2022herdphobia}, text classification \cite{chaure2019text}, and named entity recognition \cite{nie2020named}. Previous studies have also compiled relevant corpora for NLP-related tasks in the Hausa language \cite{suleiman2019towards,oyewusi2021naijaner,abubakar2021enhanced,inuwa2021first,ibrahim2022development,muhammad2022naijasenti,zandam2023online}. 
For instance, a large collection of tweets in Hausa, Igbo, Yoruba, and Nigerian Pidgin has been compiled to improve sentiment lexicons in low-resource languages \cite{abubakar2021enhanced}. Additionally, a dataset of multilingual tweets annotated with sentiment labels in major Nigerian languages has been developed \cite{muhammad2022naijasenti}. 
Despite the growing interest in low-resource languages, studies focused on offensive content detection in Hausa remain scarce. Furthermore, many languages lack sufficient linguistic resources for NLP-related tasks \cite{tsvetkov2017opportunities}. For Hausa, this scarcity is primarily due to the absence of annotated datasets and lexicons. This study proposes an approach to detect offensive online content in Hausa, one of the most widely spoken low-resource languages, thereby contributing to the broader field of computational linguistics and online safety.

\section{Methodology}
\label{sec:methodology} 
\begin{figure*}
        \centering
        \captionsetup{type=figure}
        \includegraphics[scale=0.65]{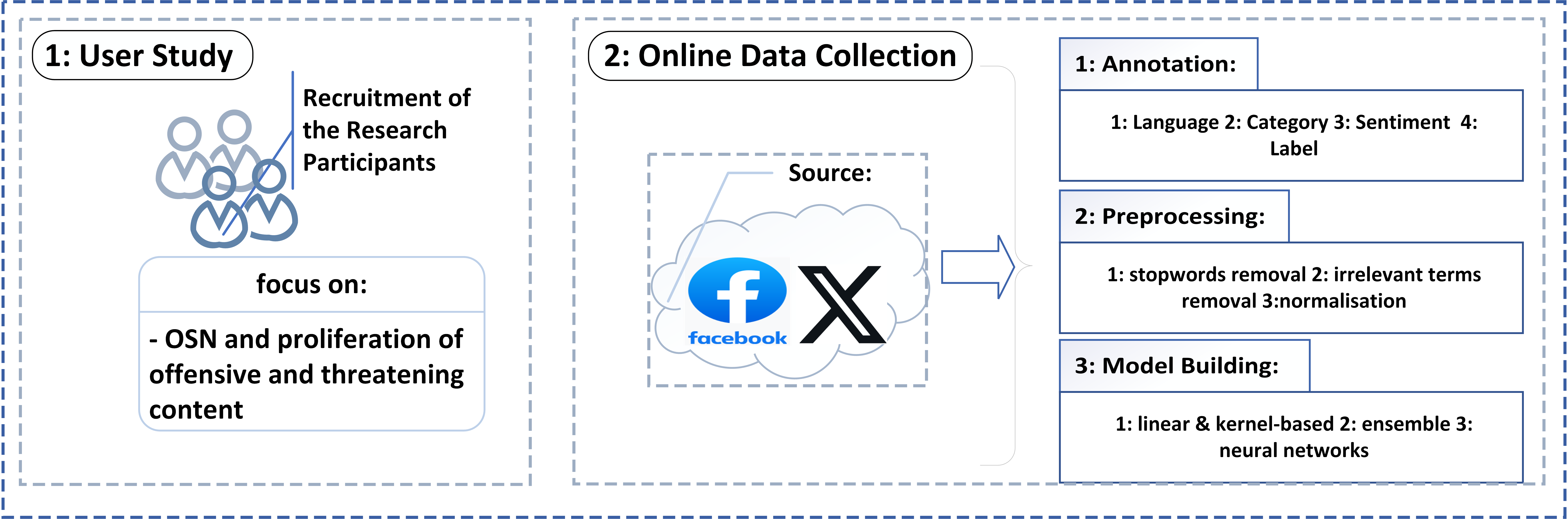}
        \caption{Overview of our approach, which involves (1) conducting user studies and (2) collecting and curating datasets to build detection systems for offensive and threatening content.}
        \label{fig:approach}
    \end{figure*}

To achieve our primary objective, we adopt the following strategy (1) engage native Hausa speakers through a user study (2) collect and annotate data comprising offensive terms in the Hausa language (3) develop and evaluate a detection system for offensive content in Hausa. Figure~\ref{fig:approach} provides an overview of our approach. 

\subsection{User Study}  
We conducted two user studies to gain insights into public perceptions of offensive and abusive online content. The study was administered via QuestionPro\footnote{\url{https://www.questionpro.com/}}, with 180 volunteers participating. The survey link was distributed on Facebook and X (formerly Twitter) to reach social media users. Informed consent was obtained from all participants, and no personally identifiable information was collected. The study was conducted in accordance with institutional ethical standards. Table~\ref{tab:demographics} presents the demographic details of the participants. The study focused on (1) the role of the Hausa language in information sharing on social media (2) the use and impact of social media as a medium for disseminating offensive and abusive content (3) public perceptions regarding the distinction between abusive terms—whether used offensively or otherwise, and (4) the different meanings and contextual interpretations of offensive and abusive content in Hausa.

    \begin{table*}[h!]
        \centering
        \captionsetup{type=table}
        \caption{Demographic information about the research participants.}  
        \label{tab:demographics}
        \renewcommand{\arraystretch}{1.2}
        \resizebox{0.9\textwidth}{!}{
        \begin{tabular}{llllllllll}
            \toprule
            &  && \textbf{User study ($n=180$) involving offensive content (HOC)} &&  &&  & \\ \hline
            
            & \textbf{Gender} && \textbf{Age} && \textbf{Education} && \textbf{Category} & \\
            \midrule
            & Female \textbf{25.3\% } && 18-25 yrs. \textbf{28\%} &&Sec. Edu. \textbf{0\%} & & Student \textbf{46\%} &\\  
            & Male \textbf{74.7\%} && 26-35 yrs. \textbf{63\%} && Undergraduate. \textbf{39\%} & & Self-employed \textbf{23\%} &\\
            &  && 36-45 yrs. \textbf{9\%} && Postgraduate \textbf{46\%} & & Civil Servant \textbf{23\%} &\\
            &  &&  &&  Graduate \textbf{15\%} && Politician \textbf{1\%}  &\\
            & &&  &&  && others \textbf{19\%} &\\ \hline
            \bottomrule
        \end{tabular}
    }
    \end{table*} 

\subsection{Dataset}
Recognising that LRLs like Hausa lack sufficient data for many NLP tasks, we collected posts from Twitter (now X) and Facebook that were likely to contain offensive or abusive content. Table~\ref{tab:data-sources} provides examples of the data sources. For data collection, we employed both an automated approach—using X's API—and a manual process for Facebook data. Algorithm \ref{alg:keywords_search} outlines the steps taken to search and retrieve relevant posts via the API. 

 \begin{algorithm}
    \small
    \caption{\emph{Text Collection}: Using keywords to collect relevant data for the study}
    \label{alg:keywords_search}
        \begin{algorithmic}[1]
        \State 
        \textbf{Initialisation:} $S = [S_i]_{i=1}^n$, list of search keywords, $n:$ number of search keywords, $M:$ number of search iterations, $N:$ number of required tweets

        \For{each iteration $m=1$ to $M$}
        \For{each query in $S$ above $i=1$ to n}
        search tweets data with keyword $S_i$ from $D=[D_j]^k_{j=1}$ where $D$ is a list of tweets 
            
        \If{$S_i$ matches $D_j$} // get relevant fields 
        \State {get posting date}
        \State get tweet ID
        \State get retweet count
        \State get favourite count 
        \State get full text 
        \State get screen-name 
        \State get urls 
    
        \Else
        \State continue 
        \EndIf
        \State {append $D_j$ to $T$ }
        \State {check for N, if satisfies maximum break} 
        \EndFor
        \EndFor
        \State {create a dataframe based on T}
        \State {\textbf{Output:} a dataframe consisting of a list of tweets $[T_t]^N_{t=1}$}
        \end{algorithmic}
    \normalsize
    \end{algorithm}

    \begin{table*}[ht]
        \centering
        \small
        \caption{Relevant keywords and sources used for data collection.} 
        \label{tab:data-sources}
        \begin{tabular}{p{2.15cm} p{11cm}}
        \hline
        \textit{Source}   & \textbf{X (formerly Twitter)}   \\
        \textit{Topics}    & politics, strike action, and social events \\
        \textit{Keywords}   & 
          siyasa (politics), ASUU strike, Maulud -Nabiyi , sallah celebration, love relationships, and zagi  \\ \hline
        \textit{Source}   & \textbf{Facebook} \\ 
        \textit{Topics}    & politics, sport, banter, and news reports \\
        \textit{Keywords}   & 
        Dandalin Siyasar Jigawa, Hakika mabudin faira, legit.ng hausa, Ayi raha asha dariya, Alfijir Hausa, Rariya, Northern hibiscus, Dokin karfe TV \\ \hline
        \end{tabular}
    \end{table*} 

To maximise the chances of obtaining offensive or abusive content, we focused on topics such as politics, sports, banter, relationships, abusive language, news reports, trending discussions, and accounts known for controversial engagements. A subset of the keywords and accounts used is listed in Table~\ref{tab:data-sources}. To further refine our dataset, we applied the following strategies for defining offensive content in Hausa: 

\begin{itemize}
    \item[-] \textit{Kamus}\footnote{Hausa dictionary \url{https://kamus.com.ng/}} (Dictionary) definition: We aligned our dataset with the standard definition of abusive language (\textit{zagi}) as provided in \textit{Kamus}. Additionally, we labelled posts based on contextual use of offensive terms, considering anecdotal evidence.
    \item[-] Past Data and User Study: We leveraged previous datasets \cite{inuwa2021first} to extract abusive words, informing our collection and annotation approach. We also incorporated offensive examples identified by participants in our user study.  
\end{itemize}

\subsubsection{Data Cleaning}

The collected dataset, especially from X, contained noise in the form of duplicates, retweets, hyperlinks, emoticons, numbers, punctuation, non-Hausa posts, and other irrelevant symbols. To ensure high-quality data for analysis, we applied the following cleaning strategies: 
\begin{itemize}
    \item[-] \textbf{Stopwords removal:} We filtered out function words that add little meaning to texts. In addition to standard stopwords, we defined a custom list for Hausa, including terms like \textit{'a', 'ni', 'to', 'su'} and non-alphabetic characters such as \textit{'.', '?', '/'}.
    \item[] \textbf{Removal of irrelevant terms:} We eliminated URLs, HTML tags, numbers, punctuation, and special symbols using regular expressions. Hashtags (\#) and mentions (@) were also removed. To filter out non-informative tweets, we calculated the number of unique words per post and discarded those with fewer than eight distinct words.
    \item[-] \textbf{Duplicate removal and normalisation:} Duplicate posts were identified and removed to prevent bias. We normalized text by converting accented vowels, replacing emojis, and correcting misspellings using a predefined Hausa dictionary. Additionally, all tokens were converted to lowercase, and stemming was applied to standardise words.    
\end{itemize} 

\subsubsection{Data Annotation}
To build an effective detection system, we manually annotated the dataset, assigning the following labels to each post: 
\begin{itemize}
    \item[-] \textit{language:} to denote the language of the post with a focus on posts in Hausa language or \textit{Engausa}\footnote{a mix of Hausa and English terms}, see Figure~\ref{fig:stats-themes-sentiment-lang}(a) for distribution. 
    \item[-] \textit{sentiment:} to indicate whether a post is positive, neutral, or negative (see Figure~\ref{fig:stats-themes-sentiment-lang}(b) for a summary). 
    \item[-] \textit{category:} to denote the theme to which the post belongs (e.g. social, political, religious, security, education, law, sport, health, agriculture, security, and business). Figure~\ref{fig:stats-themes-sentiment-lang}(c) shows the distribution of topics with the most common posts. 
    \item[-] \textit{offensive:} a column to indicate whether a post is offensive or not; abusive words are considered offensive in this regard. 
\end{itemize} 

We utilise a subset of the annotated data and incorporated into a user study, wherein feedback from online participants was used to validate the annotations, especially in instances where abusive content appeared in both playful and non-playful contexts. We compare the survey responses with the annotations to ensure consistency and reliability. 
While formal inter-rater agreement techniques are valuable for ensuring data quality, we found that incorporating survey-based ratings offered a practical means of validation under the current circumstances. We acknowledge that exposing annotators to large amounts of offensive or sensitive material raises ethical concerns. Consequently, our present ethical approval does not extend to engaging annotators with extensive offensive content without additional measures, such as mental health support and clear handling guidelines in place. Future work will integrate both formal inter-rater agreement methods and enhanced validation processes to further strengthen the work. 

\section{Experiment}
\label{sec:experiment} 
This section describes the development and evaluation of the prediction models used in detection systems. 

\subsection{Preliminary Analysis}   
\label{sec:qualitative-analysis} 
\begin{figure*}[ht]
        \centering
        \includegraphics[scale=0.45]{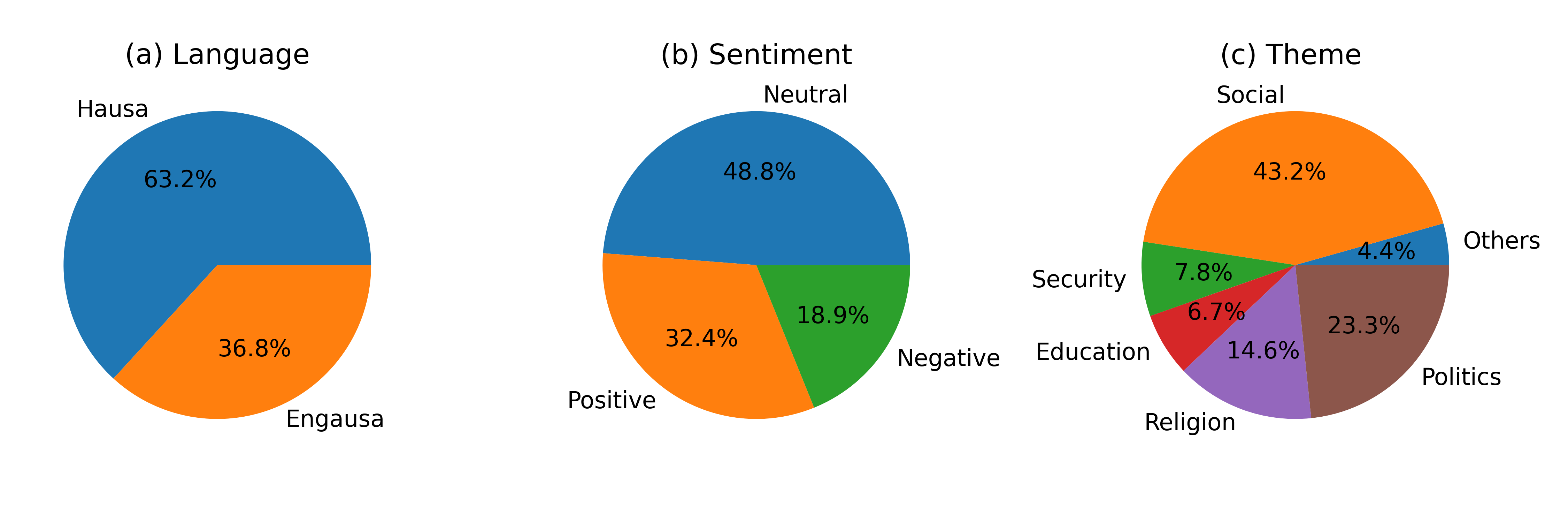}
        \caption{Statistics about language, sentiment and the major themes associated with the HOC dataset.}
        \label{fig:stats-themes-sentiment-lang}
    \end{figure*}

Figure~\ref{fig:stats-themes-sentiment-lang} provides relevant statistics about the language, sentiment and major themes in HOC collection. There are many posts written by combining Hausa and English (Engausa). The HOC attracts more diverse discussion topics (\ref{fig:stats-themes-sentiment-lang}(c)) and the low proportion of negative sentiment (\ref{fig:stats-themes-sentiment-lang}(b)) suggests that many of the abusive terms have been used in a less offensive context.  

We could not find a standard benchmark to compare how effective the detection of offensive terms would be in the Hausa language. To maximise the chances of developing an effective detection system, we explore how existing translation engines and multilingual models perform in processing Hausa terms. 
Table~\ref{tab:statistics-terms} presents the most frequent terms found in the study data. Some of these terms are quite offensive, but difficult to translate with a powerful translation engine. To determine how effective Google's translation engine performs in translating offensive Hausa words into English, we took the following top samples for comparison. 

\paragraph{Translation of Offensive Terms} Some examples of incorrectly translated offensive terms using Google's translation engine ('Hausa term' $\longrightarrow$ \textit{meaning}$\longrightarrow$ \textit{Google translation}):
    \begin{itemize}
        \item[-]  ('dan iska' $\longrightarrow$ rascal $\longrightarrow$ \textit{and iska})
        \item[-] ('yan kutumar uba' $\longrightarrow$ d***head $\longrightarrow$ \textit{father's cousins})
        \item[-] ('wawa jaki' $\longrightarrow$ idiot, donkey $\longrightarrow$ \textit{wow what})
        \item[-] ('dan jaka' $\longrightarrow$ jennet's offspring $\longrightarrow$ \textit{and jackets})
        \item[-] ('sakarai' $\longrightarrow$ dull (male) $\longrightarrow$ \textit{boy} 
        \item[-] ('sakara' $\longrightarrow$ dull (female) $\longrightarrow$ \textit{connection} 
    \end{itemize}

Despite the common use of the terms mentioned (as seen in Table~\ref{tab:statistics-terms}), translation engines, like Google's, which have access to abundant and varied data, do not perform well. In the examples given, the best results are limited to the literal meaning, and do not take into account the subtleties of the language to detect offensive contexts. This highlights the importance of enriching low-resource languages, especially in the current era of large language models (LLMs). 

    \begin{table}[ht]
    \centering
\caption{Some examples of the most frequent terms found in the data.}
\label{tab:statistics-terms}
    \begin{tabular}{llll}
        \toprule
        & \textbf{Term} & \textbf{Category} & \textbf{Proportion} \\
        \midrule
        & \textit{dan iska} & abusive & 27.7\% \\
        & \textit{kutumar uba} & abusive & 10.4\% \\ 
        & shege/shegiya & abusive & 9.0\% \\
        & \textit{jaki} & offensive & 8.7\% \\
        & \textit{uwarka} & abusive & 8.0\% \\ 
        & \textit{dan kutumar uba} & abusive & 6.9\% \\
        & \textit{wawa} & offensive & 6.9\% \\
        & \textit{jahili} & offensive & 6.90\% \\
        & \textit{ubanka} & abusive & 6.6\% \\ 
        \bottomrule
    \end{tabular}
\end{table}

\subsection{Detection System} 
The problem is modelled as a classification task consisting of two classes for both the detection tasks  - offensive/non-offensive involving the HOC data. This section describes the development and evaluation of applicable models. 

\subsubsection{Model Building} 
\label{sec:model-building} 
To determine the best detection model, we explore the following groups of machine learning models:  
\begin{itemize}
    \item[-]  \textbf{Linear and kernel-based models:} This group consists of logistic regression, a collection of regression algorithms that convey the relationship between variables (dependent and independent) and support vector machines (SVM) that categorise information separately using hyperplane to maximise the margin between them.  
    \item[-] \textbf{Naive Bayes} Naive Bayes is one of the popular models for classification tasks, especially in NLP. It is based on Bayes' theorem and assumes that features are conditionally independent given the class label. 
    \item[-] \textbf{Ensemble models:} This group comprises random forest and XGBoost models. The random forest classifier employs a set of decision trees. Closely related to the random forest is the extreme gradient boosting algorithm (XGBoost), which is based on the gradient tree boosting technique. These algorithms have been used for their fast learning and performance scalability.  
    \item[-] \textbf{Neural networks:} This group consists of multilayer perceptron (MLP) and convolutional neural network (CNN) models. The MLP is made up of an input layer, at least one hidden layer of computational neurones, and an output layer. A CNN is a deep neural network design consisting of convolutional and pooling or sub-sampling layers that feed input to a fully connected network.  
    \end{itemize} 
\paragraph{Pretrained Multilingual Models} 
In addition to the above standard ML models discussed earlier, we leverage state-of-the-art multilingual pretrained models that are widely used in NLP downstream tasks and align well with our research objectives. The models utilised in the study include: 
\begin{itemize}
    \item[-] \textbf{XLM-RoBERTa Model:} we employ the Cross-lingual Language Model - Robustly Optimised BERT Pretraining Approach  (XLM-RoBERTa) \cite{conneau2019unsupervised}, implemented via Hugging Face\footnote{\url{https://huggingface.co/FacebookAI/xlm-roberta-base}}. Trained on a large-scale multilingual dataset comprising 100 languages sourced from CommonCrawl\footnote{\url{https://commoncrawl.org/}}, this model benefits from extensive pretraining on 2.5TB of filtered data. This enables XLM-RoBERTa to capture rich cross-lingual representations, making it highly suitable for our application.
    \item[-] \textbf{BERT multilingual base model (cased):} we also utilise the multilingual Bidirectional Encoder Representations from Transformers (mBERT) model \cite{devlin2019bert}, a transformer-based architecture designed to pretrain deep bidirectional contextual representations from large-scale unlabelled text corpora. Similar to XLM-RBERTa, we use Hugging Face\footnote{\url{https://huggingface.co/google-bert/bert-base-multilingual-cased}} implementation, which has been pretrained on the top 104 languages with the largest Wikipedia presence using a masked language modelling (MLM) objective. 
\end{itemize}

\subsubsection{Feature Extraction and Training}
The process of extracting relevant training features involves converting the raw data (textual) into numerical vectors that can be fed into the learning models for prediction. 
We apply both word embedding and the Term Frequency-Inverse Document Frequency (TF-IDF) techniques in transforming the cleaned version of the data utilised in training the aforementioned classic models. Word embedding maps words to a vector space for representation. This technique enables the models to process the data and capture the semantic relationships between words \cite{mikolov2013efficient}. Using the TFIDF technique, we tried unigrams, bigrams, and trigrams to maximise the models' prediction power. 

\subsubsection{Performance Metrics} 
Performance metrics are crucial to evaluating the quality and effectiveness of machine learning models. We chose the following quantifiable metrics to assess the efficacy of the models in building the detection system:  
\begin{itemize} 
    \item[-] \textit{confusion matrix:} is a composite metric that gives the prediction output and quantification of how perplexed the model is. A true positive (TP) value signifies that the positive value is correctly predicted, a false positive (FP) means a positive value is falsely classified, a false negative (FN) means a negative value is incorrectly predicted, and a true negative (TN) means the negative value is correctly classified. 
    \item[-] \textit{accuracy:} is the number of successfully categorised instances divided by the total number of instances. This metric can be derived from the confusion matrix as the sum of TP and TN divided by the sum of TP, TN, FP and FN. 
    \item[-] \textit{precision and recall:} these are important metrics for binary classification problems.  
    \begin{itemize}
        \item Precision is the proportion of true positive instances that are classified as positive; it reflects the closeness of predicted values to one another given by: $precision = \frac{TP}{TP+FP}$. 
        \item On the other hand, recall is the proportion of positive instances that are correctly classified as positive, given by: $recall = \frac{TP}{TP+FN}$. 
        \end{itemize}
    \item[-] \textit{F1-score:} This metric combines both precision and recall into a single metric, giving equal weight to both.  
\end{itemize}
In addition to the above objective metrics, we use the Google translation tool service (see Section~\ref{sec:qualitative-analysis}) to evaluate its efficacy in translating some offensive content into the Hausa language.

\section{Results and Discussion} 
\label{sec:results-discussion} 
This study explored the issues of offensive content detection in the Hausa language through user studies and predictive analysis. In this section, we present and discuss our main findings from user studies and the detection task. 

\subsection{Public Engagement}  
\label{sec:user-study-responses}
In this section, we present and discuss the main takeaway from the set of user studies. In total, we received responses from 180 participants who participated in the surveys. 
We asked the participants to rate the degree of offensiveness (Table~\ref{tab:offensiveness-degree}) on some selected posts. 

\subsubsection{Participants' Take on Offensive Content}
According to the demographic data (Table~\ref{tab:demographics}), a typical participant was a male between 25 and 35 years old who underwent a post-graduate study. In addition to the student category, the primary work place is the public sector followed by the self-employed. A large proportion ($49\%$) of the respondents lamented the proliferation of offensive content, especially of abusive nature; $33\%$ reported encountering offensive content quite frequently. About 83\% of the respondents believe that young people tend to use abusive terms the most, and 21\% reported the likelihood of commenting on abusive online posts. 
We attribute this to the use of abusive terms in both jovial and offensive contexts. Furthermore, about 41\% of the respondents believe that abusive, hateful, or offensive content hurts online engagements. A substantial proportion (86\%) of the respondents believe that political discourse is responsible for many abusive and hateful online content.  
In Table~\ref{tab:offensiveness-degree}, about 75\% rated example 1 as very offensive. 
There are mixed ratings for example 2; about 38\% rated the post as not offensive despite the use of a term (\textit{dan iska}) that is often considered offensive. Examples 3 and 4 also received mixed ratings with 28\% and 31\% of the participants labelling them offensive. Further detail is provided in the \textit{Remarks} section of Table~\ref{tab:offensiveness-degree}. 

 \begin{table*}[ht]
        \centering
        \small
        \caption{Summary of the results from the second part of the HOC user study ($n=101$) about the public perceptions on the issue of offensive content in Hausa.} 
        \label{tab:offensiveness-degree}
        \begin{tabular}{p{2.15cm} p{11cm}}
        \hline
        \textit{Type:}   & \textbf{On the use of offensive abusive terms}   \\
        \textit{Example 1:}    & \textit{Dan gutsun uwarku watan maulid yaxo kuna bakin ciki don muna maulid amma duk tsinannan da yakarai mana happy new year sai munci uwarsa} \\
        \textit{Ratings:}   & 
        \textbf{very abusive (74.5\%);} abusive (11.2\%); somewhat abusive (7.1\%); not abusive (2\%) \\ \hline 
        
        \textit{Type:} & \textbf{On the use of subtle abusive terms}   \\
        \textit{Example 2:} & \textit{wallahi abokina basan dan iska bane sai da naga yayi 5mins yana dariya} \\
        \textit{Ratings:}   & 
        very abusive (5.1\%); abusive (4.1\%); somewhat abusive (17.3\%); might be abusive (15.3\%); not abusive (20.4\%); \textbf{not abusive at all (37.8\%)} \\ \hline 

        \textit{Type:} & \textbf{On disparaging remark}   \\
        \textit{Example 3:} & \textit{@user1 Shegiya me gudun dangi naga dae kema yar talakawa ce ko dan yanxu kin daena sae da awara da yaji iye} \\
        \textit{Ratings}   & 
        \textbf{very abusive (27.8\%); abusive (24.7\%)}; somewhat abusive (27.8\%); might be abusive (10.3\%); not abusive (5.2\%); not abusive at all (4.1\%) \\ \hline 

      \textit{Type:} & \textbf{On the use of offensive and idiomatic expression}   \\
       \textit{Example 4:} & \textit{Bari ba shegiya bace ai, Azzamula kwai,Hakkinsa kuma yana nan Wallahi ranar da DSS basu da amfani bare wannan tsinannan Mulkin naku. @user2i gara dai kasa fir'auniyar matarka ta saki dan mutane dan Wallahi yanzu muka fara bayyana Zalincin da kukama Talakawa.} \\
       \textit{Ratings:}   & 
       \textbf{very abusive (31.3\%); abusive (24\%)}; somewhat abusive (26\%); might be abusive (9.4\%); not abusive (5.2\%); not abusive at all (4.2\%) \\ \hline 
       \textbf{Remarks:}   & 
        Example 1 is offensive and involves the use of strong abusive terms. Example 2 utilises abusive term but in a playful manner. Example 3 is derogatory and Example 4 combines subtle abusive terms and idiomatic expressions to lament on the state of leadership. Most of the participants' ratings agree with these observations. \\ \hline
        \end{tabular}
    \end{table*} 

\subsection{Detection Task}
In this section, we present and discuss the efficacy of the trained models in the detection of offensive Hausa content. 

\subsubsection{Offensive Content Detection} 
\begin{figure*}[ht]
        \centering
        \includegraphics[scale=0.68]{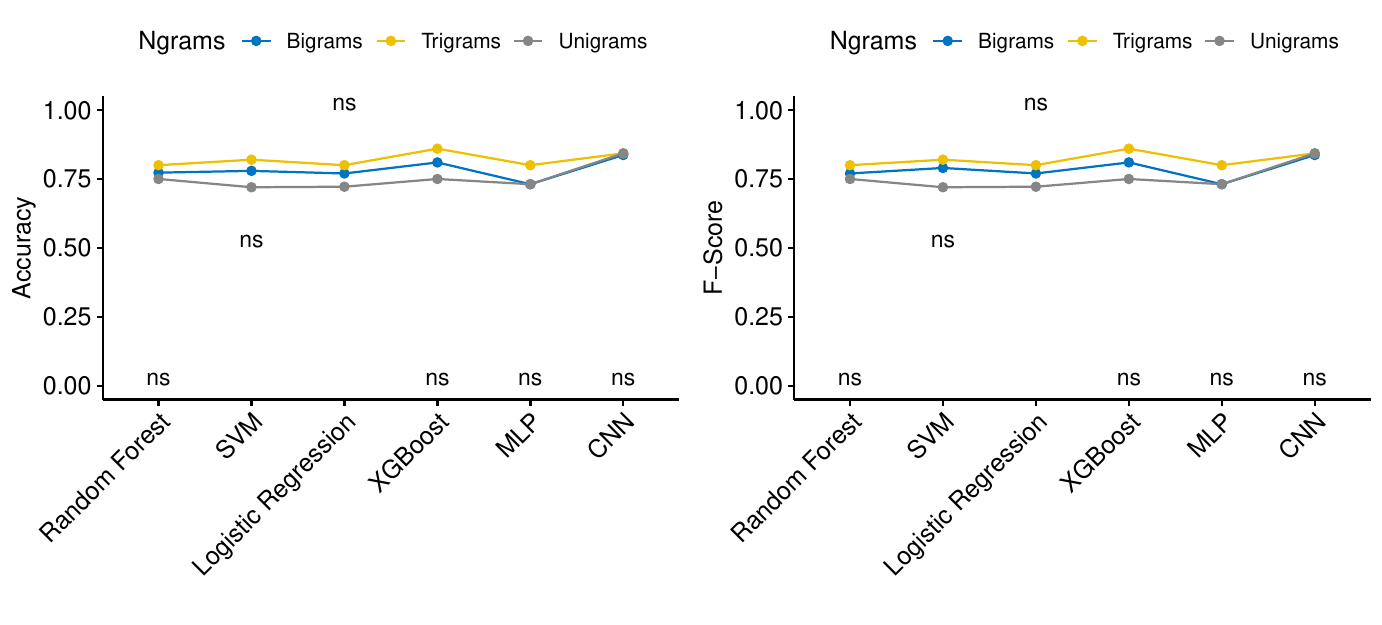}
        \caption{Accuracy and F-score performance of the trained models on the task of predicting offensive content (using the HOC dataset).}
        \label{fig:hoc_acc_fscore}
    \end{figure*}

Table~\ref{tab:hoc-unigram-bigram-trigram} and Figure~\ref{fig:hoc_acc_fscore} show the performance of trained models in the detection of online offensive content in Hausa.  XGBoost and CNN achieved the best result (with f-score values of $0.86\%$ and $0.84\%$, respectively). This is followed by the SVM, MLP, and Logistic Regression. The use of trigrams greatly improves performance in all models. These results, especially those with higher recall and precision (Tables~\ref{tab:hoc-unigram-bigram-trigram}) show a promising start in the task of detecting offensive and abusive online content in the Hausa language. Overall, the pretrained models outperform all the models. 


    \begin{table*}[ht]
        \centering
        \footnotesize
        \caption[HOC results]{Performance of the models trained on the HOC datasets for the detection of offensive content. With the exception of the CNN, the trigrams strategy results in improved performance across all the models.}
        \label{tab:hoc-unigram-bigram-trigram}
        \begin{tabular}{llp{0.5cm}p{0.7cm}p{0.5cm}p{1cm} p{0.5cm}p{0.7cm}p{0.5cm}p{1cm} p{0.5cm}p{0.7cm}p{0.5cm}p{0.7cm}p{1cm}p{0.5cm}}\hline
            & \textbf{N-grams}  & \multicolumn{4}{c}{\textbf{Unigrams}} & \multicolumn{4}{c}{\textbf{Bigrams}} & \multicolumn{4}{c}{\textbf{Trigrams}} \\   
            & \textbf{Metric} & Acc. & Recall & Prec. & F-score & Acc. & Recall & Prec. & F-score& Acc. & Recall & Prec. & F-score \\  \hline
            \multirow{9}{*}{\rotatebox{90}{\textbf{Models}}} &
            Random Forest & 0.75 & 0.75 & 0.75 & 0.75 & 0.77 & 0.77 & 0.77 & 0.77 & 0.80 & 0.80 & 0.80 & 0.80$\uparrow$ \\ 
            & SVM & 0.72 & 0.72 & 0.72 & 0.72 & 0.79 & 0.79 & 0.79 & 0.79 & 0.82& 0.82 & 0.82 & 0.82$\uparrow$ \\ 
            & Logistic Reg. & 0.72 & 0.72 & 0.72 & 0.72 & 0.77 & 0.77 & 0.77 & 0.77 & 0.80 & 0.80 & 0.80 & 0.80$\uparrow$ \\ 
            & XGBoost & 0.75 & 0.75 & 0.75 & 0.75 & 0.81 & 0.81 & 0.81 & 0.81 & \textbf{0.86} & \textbf{0.86} & \textbf{0.86} & \textbf{0.86}$\uparrow$ \\ 
            & MLP& 0.73 & 0.73 & 0.73 & 0.73 & 0.73 & 0.73 & 0.73 & 0.73 & 0.80 & 0.80 & 0.79 & 0.80$\uparrow$\\ 
            & CNN & 0.84 &  0.84 &  0.84 &  0.84 & 0.84 &  0.84 & 0.84 & 0.84  & 0.84 & 0.84  & 0.84 & 0.84\\  
            \hline
            & xlm-roberta-base & --- & --- &  --- & --- & --- &  --- & --- & ---  & 0.84 & 0.84  & 0.87 & 0.84\\  
            \hline
            & bert-base-multilingual-cased & --- & --- &  --- & --- & --- &  --- & --- & ---  & 0.86 & 0.86  & 0.89 & 0.86\\  
            \hline
        \end{tabular}
        \normalsize
    \end{table*}

N-grams are contiguous sequences of $n$ tokens in a given dataset, usually text or speech. 
To build effective detection systems, we tried various n-grams across the chosen models. We use bigrams and trigrams because of the prevalence of two- and three-adjacent words in both offensive and abusive terms. For example, \textit{dan iska} is an abusive word that is used both offensively and jovially. The word is made up of two independent words, \textit{ dan} (son of or affiliated with) and \textit{iska} (air). The use of bigrams or trigrams will maximise correct identification and prediction of the next word in a sentence. 
As demonstrated in Tables~\ref{tab:hoc-unigram-bigram-trigram},
and Figures~\ref{fig:hoc_acc_fscore} and 
the strategy of using ngrams, especially trigrams, culminated in a significant improvement in the prediction task. This has the effect of identifying common phrases and expressions in data collection.

\paragraph{Error Analysis} 
To further validate our findings, we conducted a detailed error analysis focused on identifying and examining the mistakes made by our models, particularly those with lower performance (see Table~\ref{tab:hoc-unigram-bigram-trigram}). This analysis enables us to pinpoint specific areas of difficulty and understand the underlaying reasons for suboptimal performance. Table~\ref{tab:models-error-analysis} presents examples of misclassified instances along with explanation for improving future work. 

    \begin{table*}[ht]
        \centering
        \begin{threeparttable}
        \begin{tabular}{>{\raggedright}p{2cm} >{\raggedright}p{3cm} l l l}
        \toprule
        \textbf{Models} & \textbf{Text} &  \textbf{True Class} & \textbf{Predicted Class} & \textbf{Remark} \\
        \midrule
        \textit{bert-base-multilingual-cased} & \textbf{T1} & Not Offensive & Offensive & focusing on single term \\
        \addlinespace
         & \textbf{T2} & Not Offensive & Offensive & using unknown or rare term \\
         & \textbf{T3}  & Offensive & Not Offensive & using strong abusive terms \\
        \bottomrule
        \end{tabular}
        \begin{tablenotes}
            \small
            \item \textbf{Examples:} 
            \item \textbf{T1:} \textit{Ohhh ni yar gidan mamana wallahi ina mugunjin haushin kalmar nan da wasu uwargidan ke fada wai AN AURAR MUSU MIJI Ohhh sannu me miji tinda mu zaman iskanci mukaxo ba sadaki yabayar ya auro mu ba, kofa a bude take gamasu xagi. Admin approved pls}
            \item  \textbf{T2:} \textit{Ina yan boko yaude karya ta kare muku Kuxo kufdamin sunan kanwa da turanci} 
            \item \textbf{T3:} \textit{D***N U**KA TAKA LAPIA}
            \item \textbf{T3:} \textit{??\textbf{shege dan iska} kawai abulfasadi Allah ya tsine mai albarka ????????????????punch lines paaa nie ????}
        \end{tablenotes}
        \caption{Error analysis for offensive content detection models}
        \label{tab:models-error-analysis}
        \end{threeparttable}
    \end{table*}

For the error analysis in Table~\ref{tab:models-error-analysis}, we selected the best-performing model, \textit{bert-base-multilingual-cased}, as shown in Table~\ref{tab:hoc-unigram-bigram-trigram}. We observed that many misclassifications stemmed from the model’s inability to interpret contextual meaning or its misunderstanding of certain terms. For instance, in some cases, abusive term was used within a post merely to report or describe inappropriate behaviour, yet the model labelled these posts as abusive (see \textbf{T1} in Table~\ref{tab:models-error-analysis}). This indicates that the model fails to distinguish between quoting or referencing abusive language versus actively using it. Such limitations highlight the need for more diverse and representative training data to improve contextual understanding. Moreover, the model struggles to differentiate between posts that describe abusive behaviour and those that actually exhibit it, suggesting an over-reliance on literal word presence. Another observed issue is the tendency to classify posts as abusive when they contain rare or unfamiliar words. For example, the post in \textbf{T2} uses a Hausa term \textit{kanwa}\footnote{\textit{potash} in English}, which was incorrectly flagged as abusive. Conversely, the model also failed to detect certain strongly abusive terms, underscoring its limited capacity to recognise the full range of offensive language (\textbf{T3}). 
Overall, these observations emphasise the model’s difficulty in effectively identifying abusive content in a low-resource language such as Hausa. They also underscore the importance of curating larger, higher-quality, and more representative datasets to better capture nuanced language usage. Such improvements are crucial for ensuring safer online interactions and mitigating various forms of cyberbullying.

\subsection{Discussion}

There are many similarities between the use of both offensive and abusive terms; it is not always a straightforward approach to distinguish offensive terms in the context being used. For example, in Tables~\ref{tab:statistics-terms} and ~\ref{tab:participants-comments}, there are some overlaps in the meanings associated with the terms. The distinguishing factor is the context and sentiment or polarity of the text. A high proportion of abusive terms and negative sentiment tend to point offensive content. To establish an effective distinction, manual moderation and a large collection of annotated datasets are required to discern offensive content tone. 

\paragraph{Distribution of offensive and abusive content}
To support our observation about the frequency or dominance of offensive content in political and religious discourse, we identify the proportion of each theme in the whole data (see Figure~\ref{fig:stats-themes-sentiment-lang}(c)). This prevalence can be due to the following: political and religious discourse tends to be tense and often results in physical violence or offences. Table~\ref{tab:statistics-terms} shows the most frequently associated terms with offensive and abusive content in the distribution. 

\paragraph{Topical issues vs offensive and abusive terms}
Building on the insight from the two user studies, our thematic analysis results show that political discourse, especially among young people, is at the forefront of attracting a high proportion of abusive terms (see Figure~\ref{fig:stats-themes-sentiment-lang} and Table~\ref{tab:participants-comments}). 
For offensive terms, discourse on religion, ethnicity, and political issues tops the category.  
 
    \begin{figure*}[ht]
        \centering
        \includegraphics[scale=0.78]{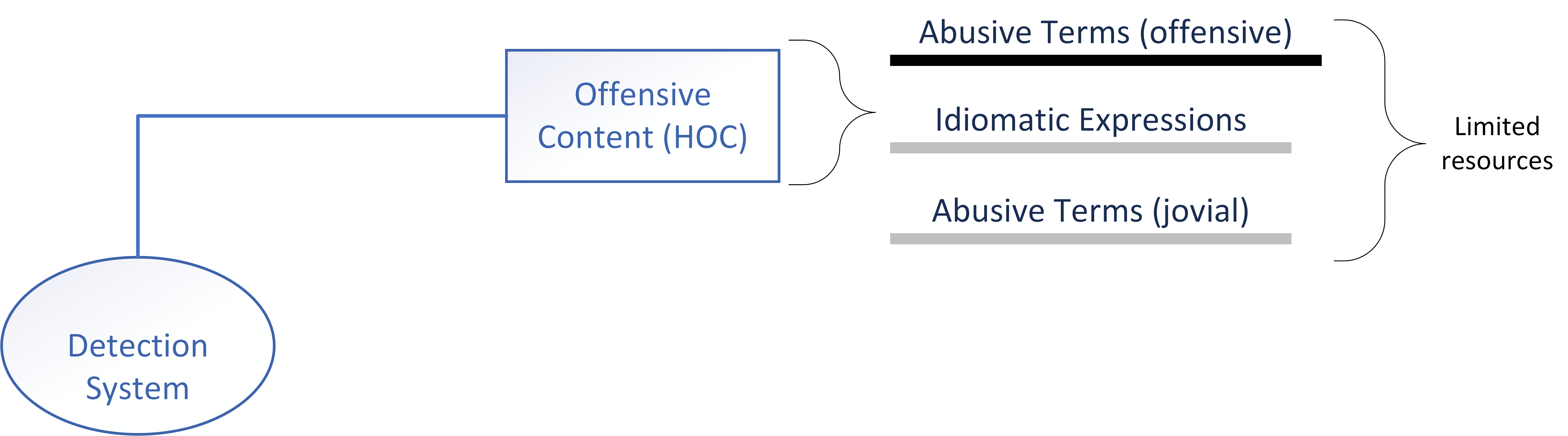}
        \caption{A summary of the areas to focus towards improving the detection system. Due to limited resources, the areas highlighted need further studies to build more effective and comprehensive detection systems for offensive content in the Hausa language.}
        \label{fig:detection-system-focus-areas}
    \end{figure*} 
    
\begin{table*}[ht]
\footnotesize
    \caption{Sample comments from the participants.}
    \label{tab:participants-comments}
    \centering
    \begin{tabular}{lp{8cm}c} \hline
    \multicolumn{1}{c}{\textbf{Type}} & \multicolumn{1}{c}{\textbf{Sample Post}} & \textbf{Category} \\ \hline
    
     & \textit{HOC Collection (Facebook and Twitter)} &  \\ \hline
        abuse & dan shegiya & offensive  \\
        abuse & \textit{Dan jaka shege dakiki} & offensive  \\ 
        abuse & dan iska  & banter  \\ 
        abuse & Dan iska da fuska kamar ichen kabari taya bazatai block naka bah, malam you need madarar tamowa fah ... & social  \\
        abuse & @user1 @user2 @user3 Nace uwarka zai gyara anan, idan baka  iya enigilishi ba kaje a fassara ma alaji.  & social  \\
        abuse & @user1 @user2 Na rantse da Allah ni nasan su si kuwa wlhi sae dae su Mutu yan kutumar uba Munafukai & political  \\
        other & @user1 At all. Ai sai an dauki at least one and a half mins ana warm ..... & other  \\ 
        abuse (allegation) &  @user1 Thank you Dr. Jeffery. Atiku is corruption personified! You need to see how he made billions as chairman of National Council on Privatization during OBJ's tenure.Barawo BANSA Atiku.I am OBIDATTI. & political \\  \hline
    \hline   
    \end{tabular}
\end{table*}

\paragraph{Distinguishing jovial and offensive abusive terms}
Ambiguity represents a semantic condition in which words or phrases can seemingly carry multiple semantic interpretations or meanings. Discerning context is crucial since the meaning derived from a given text requires many factors to be considered. The problem of ambiguity involving the Hausa language has been analysed in \cite{ishakucontrastive}. The context in which an ambiguous statement is presented is crucial in deciphering its intended meaning. 
This is required since some of the abusive terms are often used in informal and jovial conversations among friends. To make the context clear and explain the distinction, we compute the sentiment associated with each post. By extracting the abusive terms and corresponding proportions in negative and positive (and neutral) cases, the distinction is made clear. Abusive terms associated with negative sentiment often point to offensive content. 
Although offensive in tone, abusive terms are commonly used in banter and playful engagement (see Tables~\ref{tab:offensiveness-degree}, and \ref{tab:participants-comments} for examples).

\paragraph{Idiomatic expressions vs offensive content} 
Idiomatic expressions are phrases that have a figurative, non-literal meaning. They often convey a specific idea, emotion, or concept that may not be immediately apparent to someone unfamiliar with the language. Because they are often metaphorical and difficult to understand or translate for nonnative speakers, the set of words in idiomatic expressions has meanings that differ from the individual words. Some of these expressions can conceal offensive and abusive content in a more cryptic tone that will elude automated detection. To examine the interplay between idiomatic expressions conveying offensive or abusive tones, we focus on identifying the commonly used expressions and their meaning vis-a-vis offensive or abusive tones found in the data.  
Some examples of these expressions include: (1) \textit{idan maye ya manta} (2) \textit{shegiya bakar kaza}, (3) \textit{bari ba shegiya bace} and (4) \textit{abokin barawo barawo ne}. 
The above expressions can be used in different contexts and are subject to various interpretations. As shown earlier, Google's translation engine failed to correctly translate expressions (2). The bottom line is to collect a huge collection of rich data and explore various training strategies. 


\subsubsection{Impact of Offensive Terms} 
A tirade of online abuse and offense is likely to incite the public and precipitate physical violence. This is more pronounced in religious and political discourse. 
Unchecked cyberbullying could lead to continuous harassment both online and in physical space. 
When violence erupts, the use of abusive and offensive terms proliferate, thereby creating some sort of vicious cycle. At this juncture, it is pertinent to seek to understand the potential consequences of the forms of cyberbullying often encountered by online users.

\section{Conclusion}
\label{sec:conclusion} 
Online social media platforms provide users with the opportunity to express their opinions, share messages, and socialise. However, offensive and abusive content often make the online space toxic and unwelcoming. While many detection strategies have been proposed, identifying cyberbullying-related content in low-resource languages is still a challenge. With focus on the most widely spoken Chadic languages, Hausa, this study contributed the following: 
\begin{itemize}
    \item[-] The first annotated datasets consisting of offensive content to support downstream tasks in Hausa language. 
    \item[-] Detection systems to identify offensive and abusive posts in the Hausa language. 
    \item[-] Insights from two sets of user studies about the impact of offensive online content
\end{itemize} 
We found that some offensive and abusive content have subtle relationships with idiomatic expressions that are expressed in the local tone, making them difficult to detect. This underscores the need to engage native speakers and resources to better understand local conventions and demographics in order to effectively combat issues that could amount to cyberbullying in the Hausa language. 

\paragraph{Limitations and Future Work} 
Limitations of the present work include (1) lack of larger annotated datasets (2) overlapping abusive terms with ambiguous meanings, and (3) the prevalence of idiomatic expressions with subtle offensive and abusive tones. We acknowledge that our current sample of the user study is biased towards male participants. In future work, we will recruit a more diverse group of annotators to mitigate biases and enhance the fairness and representativeness of the responses. 
Moreover, the data collection was conducted as part of a postgraduate research. The student received approval to proceed after their research proposal was reviewed and endorsed by a panel of academic staff. Should further ethical review be necessary, the project is referred to the institutional review board, which undertakes additional scrutiny and provides the requisite approval. The present ethical approval does not extend to engaging annotators with extensive offensive content without additional measures, such as mental health support and clear handling guidelines in place. 
Future work will focus on enriching the data to convey nuances and better contextualisation in the language, curating relevant idiomatic expressions, and using more powerful pre-trained language models (LLMs) for fine-tuning. Engagement with diverse stakeholders, such as the general public, experts (linguistics) and major news organisations, will be necessary to improve the training data. 
Figure~\ref{fig:detection-system-focus-areas} outlines the focus areas that need to be improved. We hope this will be a starting point to motivate further research into detecting various forms of cyberbullying in low-resource languages, particularly Hausa.


\bibliographystyle{plainnat}
\bibliography{main-hoc}

\end{document}